\documentclass{article}
\usepackage{spconf,amsmath,graphicx,hyperref}
\usepackage{amsfonts}
\usepackage{amsmath,amssymb}
\usepackage{bm}
\usepackage{adjustbox}   % 只在需要时缩小表格
\usepackage{caption}  
\usepackage{booktabs,multirow}
\usepackage[table]{xcolor}   % 含 colortbl
\newcommand{\gh}[1]{\cellcolor{black!6}{#1}}  % 普通灰底
\newcommand{\ghdr}[1]{\cellcolor{black!8}{#1}}% 稍深灰底

\usepackage{float}
\usepackage[font=small]{caption}
\usepackage{tabularx}
\newcolumntype{Y}{>{\centering\arraybackslash}X} % 居中可伸缩列

\usepackage{tabularx}
\newcolumntype{Y}{>{\centering\arraybackslash}X}
\usepackage{subcaption}  % subtable
\usepackage{booktabs,multirow,xcolor,colortbl,caption}
\title{SPARSITY INDUCTION FOR ACCURATE POST-TRAINING PRUNING OF LARGE LANGUAGE MODELS}
\name{Minhao Jiang$^{1,2}$, Zhikai Li$^{1}$, Xuewen Liu$^{1,2}$,Jing Zhang$^{1,2}$,
      Mengjuan Chen$^{1,\dagger}$, Qingyi Gu$^{1,\dagger}$%
      \thanks{$^{\dagger}$ Corresponding authors.}}
\address{$^{1}$ Institute of Automation, Chinese Academy of Sciences, Beijing, China\\
         $^{2}$ School of Artificial Intelligence, University of Chinese Academy of Sciences, Beijing, China}

\begin{document}
\ninept
\maketitle
\begin{abstract}
Large language models have demonstrated capabilities in text generation, while their increasing parameter scales present challenges in computational and memory efficiency. Post-training sparsity (PTS), which reduces model cost by removing weights from dense networks, is an effective approach.
However, native dense matrices lack high sparsity, making existing approaches that directly remove weights disrupt model states, resulting in unsatisfactory performance recovery even with post-tuning.
We propose Sparsity Induction, which promotes models toward higher sparsity at both distribution and feature levels before pruning, to push the limits of PTS.
At the distribution level, we enhance distributional sparsity through mathematically equivalent scaling transformations, which are fully absorbable and incur no extra parameters or inference-time overhead.
At the feature level, we introduce Spectral Norm Loss to promote feature sparsity from a low-rank perspective.
Experiments across diverse model architectures and tasks demonstrate that our method further enhances sparsity-friendliness, achieving superior pruning performance over existing approaches.

\end{abstract}

\begin{figure*}[!t]
  \centering
  \includegraphics[width=\textwidth]{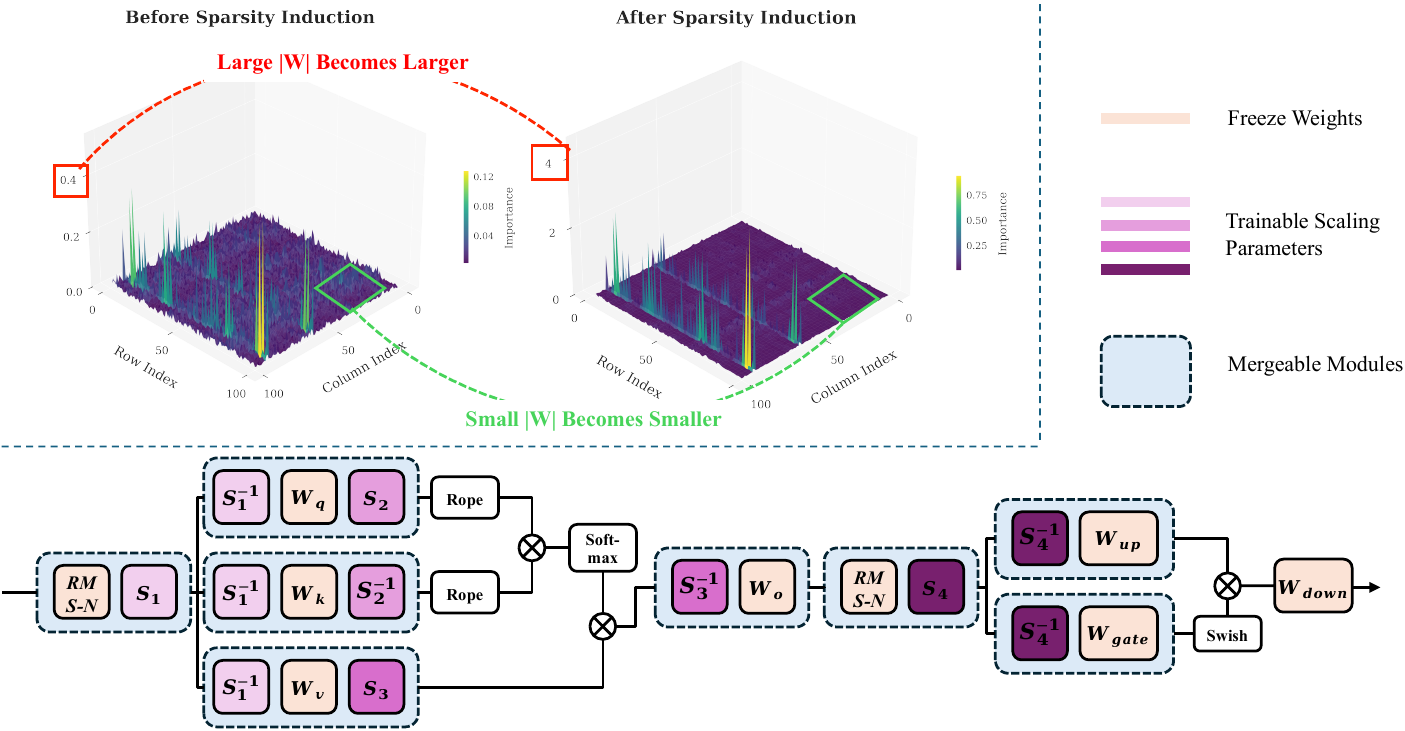}  % 或 .png/.jpg

\caption{Top: Importance distributions before (left) and after (right) Sparsity Induction . 
SI sharpens the landscape by boosting salient channels and suppressing background mass, enlarging the gap between important and unimportant weights and yielding a more sparsity-friendly distribution.
Bottom: We attach lightweight channel-wise scalers \(s\) to Attention and FFN; backbone weights are frozen and only \(s\) are trained.
Dashed modules are merged back after training, introducing no extra modules or inference-time overhead.}
   \label{fig:si-overview}
\end{figure*}

\section{Introduction}
\label{sec:intro}
Large Language Models (LLMs), such as the GPT series~\cite{C1} and the DeepSeek series~\cite{liu2024deepseek,liu2024deepseek-V2}, have achieved strong performance in language understanding and generation across diverse tasks. However, state-of-the-art LLMs typically contain tens to hundreds of billions of parameters, incurring substantial computational costs, high latency, and significant energy usage, which hinders deployment in both data centers and resource-constrained environments. Therefore, there is a pressing need for model compression methods that substantially improve inference efficiency while preserving accuracy.

Mainstream compression routes comprise pruning, quantization~\cite{shao2024omniquant,xiao2023smoothquant,li2024repquant}, and knowledge distillation~\cite{sun2019patient,sun2020contrastive}. Sparsifying LLMs is commonly discussed in terms of granularity: unstructured weight sparsity, structured patterns such as N:M~\cite{mishra2021acceleratingn:m} or block sparsity, and pruning at the level of channels~\cite{ma2023llm}, attention heads~\cite{liu2023deja}, or entire layers~\cite{sajjad2020poor}. The latter two typically alter network topology and tensor shapes, making them more intrusive to training and deployment pipelines and often necessitating substantial retraining to recover accuracy. By contrast, Post-training Sparsity (PTS ),  tends to use unstructured removal to avoid heavy retraining; it can also be combined with N:M semi-structured to obtain hardware-friendly patterns with only limited calibration and lightweight compensation, thereby delivering inference speedups and memory savings at modest overhead.

Post-training sparsity (PTS)  originally relied on magnitude pruning~\cite{frankle2018lottery}, which ranks parameters by their absolute values and zeros out small weights. This rule is easy to implement and has long served as a reference baseline for high sparsity, but in large-scale models the long-tailed magnitude distribution and inter-channel scale imbalance make a purely magnitude-based criterion prone to mispruning critical features and causing noticeable performance degradation. To mitigate this, SparseGPT~\cite{frantar2023sparsegpt} estimates loss sensitivity with a diagonal Hessian approximation under a no-retraining setting and applies row-wise compensation to redistribute the removed information to retained weights, which can greatly reduce accuracy loss even at around fifty percent sparsity. Wanda~\cite{sun2023wanda} further introduces the L2 norm of input activations as a data-aware criterion, yielding more robust pruning decisions at very low additional cost. Recent work~\cite{dong2024pruner,das2023beyond} has broadened the sources and combinations of auxiliary signals and leveraged symbolic regression to automatically search importance expressions over weights and activations, aiming for stable recovery at higher sparsity levels. Nevertheless, the returns from merely strengthening importance scores have become marginal, indicating that post-training sparsity should be reconsidered from more fundamental perspectives such as preserving distributional and scaling consistency.

Across existing post-training sparsity methods, pushing for ever more refined importance scores exhibits diminishing returns. These scores are typically built on local approximations or limited calibration data; improvements increasingly yield marginal re-ranking, while the added computation and tuning accumulate and fail to translate into stable accuracy gains at large scale and high sparsity. 
As shown by the left panel of Fig.~\ref{fig:si-overview} (top), the baseline importance landscape indicates that the raw weight distribution is not inherently sparse; naive pruning therefore removes useful mass and perturbs activations. 
More fundamentally, pruning disrupts the self-consistent distributional and scaling structure established during training, including the statistical shape of weights and the relative scales across channels. The resulting imbalance propagates through attention and feed-forward paths and emerges as the primary driver of accuracy loss in post-training sparsity. Merely refining importance scores cannot remedy this distributional damage. A more promising direction is to preserve or rapidly recalibrate distribution and scale consistency before and after pruning, retaining the low-overhead nature of post-training sparsity while suppressing error amplification at high sparsity levels.

We introduce the central idea of Sparsity Induction (SI): before pruning, we proactively shape the model’s weight distribution and feature structure to make it more sparsity-friendly, thereby improving the accuracy and stability of PTS without adding inference overhead. To instantiate this idea, we apply absorbable, mathematically equivalent scaling transformations at the distributional level, using a few learnable scale factors to align channel/head statistics and folding them entirely into the parameters at inference; at the feature level, we add a spectral-norm loss to encourage low-rank structure and promote feature sparsity. We further devise an efficient optimization framework that jointly updates standard importance scores and scaling factors, yielding up to ~20× speedup over conventional pipelines. Extensive experiments across architectures and tasks show that SI markedly strengthens sparsity-friendliness and consistently surpasses existing approaches in pruning performance.

Our contributions can be summarized as follows:

\textbf{·} We introduce the concept of sparsity induction, which enforces sparsity in the original model before pruning.

\textbf{·} We design a sparsity induction framework that optimizes sparsity properties from both distributional and feature-level perspectives.

\textbf{·} Extensive experiments demonstrate that our approach substantially enhances sparsity-friendliness and significantly improves pruning performance.

\section{METHODOLOGY}
\label{sec:format}

\subsection{Preliminaries}
\label{ssec:subhead}
Most pruning pipelines construct a binary mask according to a sparsity metric and a target sparsity level, and obtain a pruned weight matrix by zeroing the masked entries. Let \(\mathbf{W}\in\mathbb{R}^{d_{\mathrm{out}}\times d_{\mathrm{in}}}\) denote a dense weight matrix and \(\mathbf{M}\in\{0,1\}^{d_{\mathrm{out}}\times d_{\mathrm{in}}}\) the pruning mask; the pruned weights are given as follows,
\begin{equation}
\widehat{\mathbf{W}} = \mathbf{W} \odot \mathbf{M} \text{,}
\label{eq:masked-weight}
\end{equation}
where \(\odot\) denotes the Hadamard product. For an input vector \(x\in\mathbb{R}^{d_{\mathrm{in}}}\), the output distortion induced by pruning is given as follows,
\begin{align}
\Delta Y &= (\mathbf{W} - \widehat{\mathbf{W}})\,X \text{,}
\label{eq:output-distortion}
\end{align}
where \(\Delta Y\) denotes the pruning-induced output distortion.

Existing methods primarily focus on designing effective metrics so that the induced mask \(\mathbf{M}\) minimizes such distortion. However, the original dense weight matrices in LLMs typically do not exhibit substantial natural sparsity. In this setting, the pruned model \(\widehat{\mathbf{W}}\) is a strict subset (hard-masked version) of \(\mathbf{W}\); without adapting the weight distribution or the feature responses, na\"{\i}ve hard masking often yields nontrivial performance degradation.

In this work, we induce sparsity adaptively from two complementary perspectives, distributional and feature, to prepare the model for pruning: (i) we reshape the weight distribution to enhance separability between important and unimportant parameters; and (ii) we learn masks under feature-aware objectives that directly penalize output distortion. This dual guidance sparsifiable structure and improves the performance of the sparse model.

\subsection{Sparsity Induction: Distributions}
\label{ssec:subhead}

We aim to perform sparsity induction so that, prior to pruning, the weight distribution is pre-adapted to the accuracy degradation induced by metric-based pruning, thereby effectively enhancing sparsity. However, for high-dimensional LLMs it is difficult to determine the correct direction of distributional optimization by mere inspection, while full-parameter fine-tuning is prohibitively resource-intensive. To achieve precise yet efficient control, we introduce a functionally equivalent reparameterization based on channel-wise scaling and channel-wise shifting, which reshapes the weight distributions without altering model functionality.This equivalent transformation can be applied to both the linear layer and attention operation.

\textbf{Linear layer.}
To reshape weight statistics without changing functionality, we introduce per-channel
scaling $\boldsymbol{s}>0$ and per-channel shifting $\boldsymbol{\delta}$, and let $\mathbf{S}$ be the diagonal matrix of $\boldsymbol{s}$
(i.e., $\mathbf{S}=\mathrm{diag}(\boldsymbol{s})$). 
For a linear layer $\mathbf{Y}=\mathbf{W}\mathbf{X}+\mathbf{b}$, with per-channel scale $\boldsymbol{s}>0$,
shift $\boldsymbol{\delta}$, and $\mathbf{S}=\mathrm{diag}(\boldsymbol{s})$, we can therefore express the equivalent reparameterization as follows,
\begin{equation}
\mathbf{Y} \;=\; \mathbf{W}\mathbf{X}+\mathbf{b} \;=\;
\underbrace{\mathbf{S}^{-1}(\mathbf{X}-\boldsymbol{\delta})}_{\tilde{\mathbf{X}}}\;
\underbrace{(\mathbf{W}\mathbf{S})}_{\tilde{\mathbf{W}}}
\;+\;
\underbrace{(\mathbf{b}+\mathbf{W}\boldsymbol{\delta})}_{\tilde{\mathbf{b}}} \text{,}
\label{eq:equiv-linear}
\end{equation}
where $\tilde{\mathbf{X}}=\mathbf{S}^{-1}(\mathbf{X}-\boldsymbol{\delta})$, $\tilde{\mathbf{W}}=\mathbf{W}\mathbf{S}$, and $\tilde{\mathbf{b}}=\mathbf{b}+\mathbf{W}\boldsymbol{\delta}$; $\mathbf{S}=\mathrm{diag}(\boldsymbol{s})$ is a diagonal matrix, and $\mathbf{W}$ is the (dense) weight matrix.

\textbf{Attention.}
To keep attention logits intact (thus preserving the softmax attention and outputs),
we apply a pair of inverse per-channel scalings to queries and keys: with $\boldsymbol{s}_a>0$
and $\mathbf{S}_a=\mathrm{diag}(\boldsymbol{s}_a)$, set $\tilde{\mathbf{Q}}=\mathbf{Q}\mathbf{S}_a$ and $\tilde{\mathbf{K}}=\mathbf{K}\mathbf{S}_a^{-1}$;
then $\tilde{\mathbf{Q}}\tilde{\mathbf{K}}^\top = \mathbf{Q}\mathbf{K}^\top$.
The factors can be absorbed into the $\mathbf{Q}/\mathbf{K}$ projection weights, reshaping their
distributions without altering functionality.

\textbf{Lightweight objective.}
We learn only a small set of transformation parameters on calibration data
(e.g., $\boldsymbol{s},\boldsymbol{\delta}$ for linear layers and $\boldsymbol{s}_a$ for attention) using a lightweight
objective that pre-adapts the model to the mask. Accordingly, the optimization objective is given as follows,
\begin{equation}
\min_{\Theta}\ \mathbb{E}_{\mathbf{X}}\,\|(\tilde{\mathbf{W}}-\tilde{\mathbf{W}}\odot \mathbf{M})\tilde{\mathbf{X}}\|_2^2,
\qquad \Theta=\{\boldsymbol{s},\boldsymbol{\delta},\boldsymbol{s}_a\} \text{,}
\label{eq:preadapt-final}
\end{equation}

where $\Theta$ denotes the transformation parameters, $\mathbf{M}$ is a binary mask meeting the target sparsity, and $\tilde{\mathbf{W}}$ and $\tilde{x}$ follow the definitions above, as shown in the bottom panel of Fig.~\ref{fig:si-overview}.
This avoids full-parameter fine-tuning while    improving mask-friendliness.

\subsection{Sparsity Induction: Features}
\label{ssec:subhead}

After the distributional reparameterization, optimization at the feature level may still lack a stable and well-defined search direction. We address this by combining a data-driven initialization of the transformation factors with a spectral-style guidance term so as to obtain faster and more reliable convergence while keeping the number of trainable parameters small. Concretely, we construct a robust channel-wise initialization as follows,
Let \(X=\{\mathbf{x}_j\}_{j=1}^{n}\) denote the calibration set and \(\mathbf{y}_j=\mathbf{W}\mathbf{x}_j\) the pre-transform outputs of a layer. For output channel \(i\), define the robust statistics \(m_i=\operatorname{median}_{j}\,\mathbf{y}_{j,i}\) and \(\mu_i=\tfrac{1}{n}\sum_{j=1}^{n}\mathbf{y}_{j,i}\); we then set
\begin{equation}
s_i^{(0)}=\bigl(m_i - g(\mu_i)\bigr)^2 \text{,}
\label{eq:init-scale}
\end{equation}
where \(g:\mathbb{R}\!\to\!\mathbb{R}\) is a fixed monotone mapping (e.g., identity or simple rescaling), and \(s_i^{(0)}\) denotes the initial per-channel scale.

On top of this initialization, we optimize only the transformation factors using an output-matching loss combined with a spectral-style regularizer; the objective is defined as follows,
\begin{equation}
\mathcal{L}_{\text{total}}=\mathcal{L}_{\text{MSE}}+\lambda\,\mathcal{R}_{p} \text{,}
\label{eq:total-loss}
\end{equation}
\begin{equation}
\mathcal{L}_{\text{MSE}}=\operatorname{MSE}\!\bigl(\mathbf{Y}^{\mathrm{d}},\,\mathbf{Y}^{\mathrm{s}}\bigr) \text{,}
\label{eq:mse}
\end{equation}
\begin{equation}
\mathcal{R}_{p}=\exp\!\Bigl(-\alpha\sum_{\ell\in\mathcal{S}}\|\mathbf{W}_{\ell}\|_{p}\Bigr) \text{,}
\label{eq:pnorm}
\end{equation}
where \(\mathbf{Y}^{\mathrm{d}}\) and \(\mathbf{Y}^{\mathrm{s}}\) denote the dense and sparsity-induced batched outputs, respectively; \(\lambda>0\) balances reconstruction and regularization; \(\alpha>0\) controls the regularizer strength; \(p\!\ge\!1\) is the norm order; \(\mathcal{S}\) indexes the set of layers subject to sparsity induction; and \(\mathbf{W}_{\ell}\) is the weight matrix of layer \(\ell\).
The combination of robust initialization and spectral guidance narrows the search space, stabilizes feature-level updates, and improves pruning accuracy at a fixed sparsity without resorting to full-parameter fine-tuning. 
\subsection{Efficiency: Fast Hessian Update}
We avoid repeated forwards by replacing activation variance with a diagonal Gauss--Newton/Hessian proxy for a linear layer $y=\mathbf{W}\mathbf{X}+\mathbf{b}$; the proxy is defined as follows,
\begin{equation}
\mathbf{H}\ \propto\ {\Sigma}_{\mathbf{X}}\ \triangleq\ \mathbb{E}[\mathbf{X}\mathbf{X}^\top] \text{,}
\label{eq:proxy-h}
\end{equation}
where $\mathbf{H}$ serves as a curvature proxy and is diagonal up to a positive layer-wise constant, and $\boldsymbol{\Sigma}_{\mathbf{X}}$ denotes the input covariance.

When distributional scaling is folded into parameters at inference, with $\mathbf{X}'=\mathbf{D}\mathbf{X}$ and $\mathbf{D}=\mathrm{Diag}(\boldsymbol{s})$, the proxy transforms as follows,
\begin{equation}
\mathbf{H}'\ \propto\ \mathbf{D}\,\mathbf{H}\,\mathbf{D}
\quad\Longrightarrow\quad
\mathrm{diag}(\mathbf{H}')\ =\ \boldsymbol{s}^{2}\!\circ\mathrm{diag}(\mathbf{H}) \text{,}
\label{eq:h-transform}
\end{equation}
where $\boldsymbol{s}\in\mathbb{R}^{d_{\text{in}}}$ is the absorbable per-channel scaling vector, $\mathbf{D}=\mathrm{Diag}(\boldsymbol{s})$ is its diagonal embedding, and $\circ$ denotes the Hadamard product.

Substituting the proxy gives the constant-time refreshable score as follows,
\begin{equation}
m_{\mathrm{fast}}(\mathbf{W})
\ \triangleq\
|\mathbf{W}|\circ \sqrt{\mathrm{diag}(\mathbf{H}') }
\ \propto\
|\mathbf{W}|\circ \bigl(\sqrt{\mathrm{diag}(\mathbf{H})}\circ \boldsymbol{s}\bigr) \text{,}
\label{eq:wanda-fast}
\end{equation}
where $|\cdot|$ and $\mathrm{diag}(\cdot)$ denote the elementwise absolute value and diagonal extraction, respectively. By \eqref{eq:h-transform}, $m_{\mathrm{fast}}(\mathbf{W})$ induces the same ordering as the classical Wanda metric evaluated on the scaled input $\mathbf{X}'=\mathbf{D}\mathbf{X}$ up to a positive layer-wise constant, thus preserving rankings while enabling $O(d_{\text{in}})$ refresh via cached diagonals. This algorithmic design greatly accelerates our overall method.

\section{EXPERIMENTS}
\label{tab:zs-llama12}
\subsection{Experiment Setup}
% \subsubsection{Model, Dataset and Evaluation}
\textbf{Models, Datasets, and Evaluations}
 We evaluate eight decoder-only LLMs that jointly cover size and architectural style: OPT--125M/350M/1.3B/2.7B~\cite{zhang2022opt} and LLaMA-1/2~\cite{touvron2023llama,touvron2023llama2} at 7B and 13B. This set spans the bias-heavy OPT family and the largely bias-light LLaMA family, enabling representative comparisons across structures and capacities. For language modeling, we report perplexity on the C4~\cite{c4} and WikiText-2~\cite{wiki} validation sets. To assess downstream generalization, we measure zero-shot accuracy with EleutherAI LM Harness on six tasks: ARC-Easy, ARC-Challenge~\cite{clark2018thinkARC}, HellaSwag~\cite{zellers2019hellaswag}, BoolQ~\cite{clark2019boolq}, Winogrande~\cite{sakaguchi2021winogrande}, and OpenBookQA~\cite{mihaylov2018canopenbookQA}. Results are compared against the dense models and against pruning baselines under identical tokenization, prompts, and scoring; calibration data are disjoint from all evaluation sets.

\textbf{Details.}
 Pruning is applied to the linear layers in attention and MLP blocks; embeddings and the final LM head remain dense, and biases are not pruned. We compare three post-training sparsification baselines---Magnitude, Wanda, and SparseGPT---and employ a plug-and-play sparsity-induction step compatible with each baseline to stabilize the selected masks/thresholds. Following prior work, we use exactly \textbf{128 activation samples} (2048-token segments from the C4 training set), shared across all models and methods. All experiments use PyTorch and HuggingFace Transformers with LM Harness on a single \textbf{NVIDIA RTX A6000} (48\,GB). Baseline pruning is one-shot without task-specific fine-tuning; the induction stage performs a short \textbf{1--5 epoch} pass on the 128 samples. We use a maximum context length of 2048 tokens and adjust batch sizes to fit memory.

\subsection{Experiment Results and Analysis}

Table~\ref{tab:opt-llama1-wt2-ppl} reports perplexity for both dense and sparse models, with and without SI augmentation. SI systematically mitigates the perplexity degradation introduced by pruning, with the largest gains under aggressive sparsity where baseline sparse models otherwise suffer severe quality loss or become practically unusable.

Table~\ref{tab:zs-llama12} summarizes the average zero-shot performance of dense models and their pruned counterparts across six benchmarks.Across multiple sparsity regimes, adding Sparsity Induction (SI) consistently improves the zero-shot accuracy of one-shot pruning baselines, including magnitude pruning, Wanda, and SparseGPT.

Although SI is not explicitly tailored to N:M semi-structured sparsity, it nevertheless delivers strong results under such patterns. Because SI’s auxiliary parameters can be folded into the weight matrices before inference, it introduces no additional runtime modules or overhead. Consequently, models equipped with SI remain fully compatible with hardware acceleration for N:M sparsity and can be deployed for speedups without modification.

As summarized in Table~\ref{tab:fast-vs-classic}, our fast Hessian update reduces the Wanda refresh time on LLaMA-7B (batch size $=1$) from $345.91$\,s to $15.27$\,s, a $22.65\times$ improvement.
For end-to-end latency under the $2{:}4$ sparsity pattern (Table~\ref{tab:e2e-nm24}), we observe $251$\,ms versus $312$\,ms on LLaMA-7B ($1.24\times$ speedup).
Because SI is fully absorbable, it introduces no additional runtime overhead relative to Wanda or SparseGPT.

% ===== 单栏表格：OPT + LLaMA-1 perplexity on WikiText-2 =====
\begin{table}[H]
\centering
\begingroup
\scriptsize
\setlength{\tabcolsep}{3.2pt}
\renewcommand{\arraystretch}{1.08}
\captionsetup{font=small}

\caption{Perplexity $\downarrow$ on the WikiText-2 dataset for OPT and LLaMA-1 models.
Bold indicates the lowest PPL within each (Sparsity, Model) block.
\textbf{Abbrev.:} \textbf{SI} = \emph{Sparsity Induction}; rows marked “\textbf{+SI}” mean the base method + SI.}
\label{tab:opt-llama1-wt2-ppl}

{\arrayrulecolor{black}%
\begin{tabularx}{\columnwidth}{@{}c l *{6}{Y}@{}}
\toprule
& & \multicolumn{4}{c}{\textbf{OPT}} & \multicolumn{2}{c}{\textbf{LLaMA-1}} \\
\cmidrule(lr){3-6}\cmidrule(lr){7-8}
\textbf{Sparsity} & \textbf{Method} & \textbf{125M} & \textbf{350M} & \textbf{1.3B} & \textbf{2.7B} & \textbf{7B} & \textbf{13B} \\
\midrule
0\% & \ghdr{Dense} & \ghdr{27.65} & \ghdr{22.00} & \ghdr{14.62} & \ghdr{12.47} & \ghdr{5.68} & \ghdr{5.09} \\
\midrule

\multirow[c]{6}{*}{\cellcolor{white}\textbf{50\%}}
& Magnitude        & 193.36 & 97.78 & 1712.82 & 265.21 & 17.28 & 20.21 \\
& \gh{+SI (Ours)}  & \gh{52.85} & \gh{49.55} & \gh{26.39} & \gh{18.86} & \gh{12.08} & \gh{18.45} \\
& Wanda            & 38.99 & 36.19 & 18.40 & 14.22 & 7.26 & 6.15 \\
& \gh{+SI (Ours)}  & \gh{38.00} & \gh{34.78} & \gh{18.27} & \gh{14.19} & \gh{\textbf{7.04}} & \gh{\textbf{6.04}} \\
& SparseGPT        & 36.97 & 31.40 & \textbf{17.40} & 13.46 & 7.17 & 6.22 \\
& \gh{+SI (Ours)}  & \gh{\textbf{35.79}} & \gh{\textbf{31.30}} & \gh{17.45} & \gh{\textbf{13.45}} & \gh{7.13} & \gh{6.12} \\
\midrule

\multirow[c]{6}{*}{\cellcolor{white}\textbf{2{:}4}}
& Magnitude        & 341.45 & 417.04 & 427.18 & 1153.12 & 42.54 & 18.36 \\
& \gh{+SI (Ours)}  & \gh{226.83} & \gh{173.99} & \gh{40.73} & \gh{31.14} & \gh{18.55} & \gh{15.55} \\
& Wanda            & 79.89 & 112.57 & 28.16 & 21.25 & 11.53 & 9.60 \\
& \gh{+SI (Ours)}  & \gh{79.11} & \gh{94.80} & \gh{27.54} & \gh{20.23} & \gh{\textbf{10.51}} & \gh{\textbf{8.32}} \\
& SparseGPT        & 61.44 & 49.55 & 23.87 & 17.10 & 11.00 & 9.05 \\
& \gh{+SI (Ours)}  & \gh{\textbf{60.66}} & \gh{\textbf{49.51}} & \gh{\textbf{23.85}} & \gh{\textbf{17.07}} & \gh{10.65} & \gh{8.78} \\
\midrule

\multirow[c]{6}{*}{\cellcolor{white}\textbf{4{:}8}}
& Magnitude        & 169.09 & 160.73 & 240.15 & 166.94 & 16.83 & 13.87 \\
& \gh{+SI (Ours)}  & \gh{122.89} & \gh{74.17} & \gh{28.20} & \gh{21.44} & \gh{13.97} & \gh{13.25} \\
& Wanda            & 53.17 & 59.12 & 22.17 & 16.57 & 8.58 & 7.40 \\
& \gh{+SI (Ours)}  & \gh{53.06} & \gh{54.30} & \gh{21.88} & \gh{16.47} & \gh{\textbf{8.15}} & \gh{\textbf{6.90}} \\
& SparseGPT        & 44.81 & 39.20 & 20.25 & 14.98 & 8.56 & 7.43 \\
& \gh{+SI (Ours)}  & \gh{\textbf{43.63}} & \gh{\textbf{39.11}} & \gh{\textbf{20.03}} & \gh{\textbf{14.93}} & \gh{8.35} & \gh{7.26} \\
\bottomrule
\end{tabularx}%
}
\endgroup
\end{table}
% ==== 单栏表格 ====
\begin{table}[H]
\centering
\begingroup
\scriptsize                               % 9pt，不小于9pt
\setlength{\tabcolsep}{3.2pt}
\renewcommand{\arraystretch}{1.08}
\captionsetup{font=small}

\caption{Average zero-shot accuracy (\%) $\uparrow$ on ARC-Easy, ARC-Challenge, HellaSwag, BoolQ, Winogrande, and OpenBookQA; models are LLaMA-1/2 at 7B and 13B.}

\label{tab:zs-llama12}

{\arrayrulecolor{black}%
\begin{tabularx}{\columnwidth}{@{}c l *{4}{Y}@{}}
\toprule
 & & \multicolumn{2}{c}{\textbf{LLaMA-1}} & \multicolumn{2}{c}{\textbf{LLaMA-2}} \\
\cmidrule(lr){3-4}\cmidrule(lr){5-6}
\textbf{Sparsity} & \textbf{Method} & \textbf{7B} & \textbf{13B} & \textbf{7B} & \textbf{13B} \\
\midrule
0\% & \ghdr{Dense} & \ghdr{54.79} & \ghdr{57.82} & \ghdr{55.95} & \ghdr{58.25} \\
\midrule

\multirow[c]{6}{*}{\cellcolor{white}\textbf{50\%}}
& Magnitude        & 43.17 & 46.27 & 48.04 & 53.40 \\
& \gh{+SI (Ours)}         & \gh{47.52} & \gh{46.85} & \gh{48.68} & \gh{54.27} \\
& Wanda            & 51.47 & \textbf{55.45} & 53.63 & 56.10 \\
& \gh{+SI (Ours)}         & \gh{\textbf{52.50}} & \gh{55.32} & \gh{\textbf{53.91}} & \gh{56.08} \\
& SparseGPT        & 52.37 & 54.31 & 53.34 & 55.71 \\
& \gh{+SI (Ours)}         & \gh{52.04} & \gh{54.16} & \gh{53.10} & \gh{\textbf{56.22}} \\
\midrule

\multirow[c]{6}{*}{\cellcolor{white}\textbf{2{:}4}}
& Magnitude        & 43.55 & 46.02 & 46.01 & 47.84 \\
& \gh{+SI (Ours)}         & \gh{42.66} & \gh{47.28} & \gh{46.39} & \gh{51.12} \\
& Wanda            & 45.75 & 48.73 & 45.95 & 50.06 \\
& \gh{+SI (Ours)}         & \gh{45.71} & \gh{49.54} & \gh{46.32} & \gh{50.98} \\
& SparseGPT        & 46.07 & 49.31 & \textbf{47.23} & 52.14 \\
& \gh{+SI (Ours)}         & \gh{\textbf{46.30}} & \gh{\textbf{50.05}} & \gh{47.04} & \gh{\textbf{52.59}} \\
\midrule

\multirow[c]{6}{*}{\cellcolor{white}\textbf{4{:}8}}
& Magnitude        & 46.18 & 48.46 & 48.77 & 52.10 \\
& \gh{+SI (Ours)}         & \gh{46.07} & \gh{49.87} & \gh{49.17} & \gh{54.49} \\
& Wanda            & 48.58 & 52.22 & 50.02 & 54.59 \\
& \gh{+SI (Ours)}         & \gh{48.89} & \gh{\textbf{52.61}} & \gh{50.37} & \gh{54.79} \\
& SparseGPT        & 48.73 & 51.78 & 50.48 & 55.09 \\
& \gh{+SI (Ours)}         & \gh{\textbf{49.05}} & \gh{52.54} & \gh{\textbf{50.90}} & \gh{\textbf{55.10}} \\
\bottomrule
\end{tabularx}
}
\endgroup
\end{table}

\vspace{-4pt} 
% ===================== Table 3 =====================
\begin{table}[H]
\centering
\scriptsize 
\caption{Fast Hessian update vs.\ classical activation-recompute for refreshing the Wanda metric on LLaMA-7B in a epoch of 128 samples(batch size = 1).}
\label{tab:fast-vs-classic}
\begin{tabular}{lccc}
\toprule
Method & Update time (s) & Avg. time/iter (s) & Speedup \\
\midrule
Classical Recompute & 345.91 & 2.70 & $1.00\times$ \\
% Fast Update (Ours)  & {\textbf{15.27}  & {\textbf{0.12} & ${\textbf{22.65}\times$ \\
Fast Update (Ours) & \textbf{15.27} & \textbf{0.12} & $\textbf{22.65}\times$ \\
\bottomrule
\end{tabular}
\end{table}

\vspace{-4pt} 
% ===================== Table 4 =====================
\begin{table}[H]
\centering
\scriptsize 
\caption{End-to-end latency under $2{:}4$ sparsity on LLaMA-7B. Our SI is fully absorbable and adds \emph{zero} runtime latency compared with wanda.}
\label{tab:e2e-nm24}
\begin{tabular}{lcccc}
\toprule
 Pattern  & E2E latency (ms) & Speedup \\
\midrule
  Dense & 312 & $1.00\times$ \\
$2{:}4$  (Wanda)      & 251 & $1.24\times$ \\
$2{:}4$  (Wanda+SI)      & 251 & $1.24\times$ \\
\bottomrule
\end{tabular}
\end{table}

\section{CONCLUSIONS}
\label{sec:typestyle}
We have provided Sparsity Induction (SI), a training-light framework that prepares large language models for pruning. SI acts along two complementary dimensions, distributional and feature level, by applying functionally equivalent per channel transformations consisting of scaling and shifting, and by learning only a small set of parameters on calibration data. This preadaptation suppresses pruning-induced output distortion and reduces accuracy loss while avoiding full parameter fine tuning.

Experiments across a range of model sizes and architectures, and with multiple PTS methods, show that SI serves as an easily integrated enhancement that improves the performance of sparse models. The results further indicate that per channel scaling and shifting provide an effective mechanism for restoring expressivity after masking. Future work will explore richer equivalence transformations and structure-aware variants, as well as tighter integration with quantization and other sparsification techniques, in order to further advance sparse large language model performance.

\section{ACKNOWLEDGEMENTS}
This work was supported in part bythe Strategic Priority Research Program of the Chinese Academy of Sciences (Grant No. XDB1100000); in part by the National Natural Science Foundation of China under Grant Number 62276255; in part by the Postdoctoral Fellowship Program of CPSF under Grant Number GZC20251175; 

\bibliographystyle{IEEEbib}
\bibliography{strings,refs}

@article{C1,
  title={Gpt-4: A review on advancements and opportunities in natural language processing},
  author={Baktash, Jawid Ahmad and Dawodi, Mursal},
  journal={arXiv preprint arXiv:2305.03195},
  year={2023}
}

@article{liu2024deepseek,
  title={Deepseek-v3 technical report},
  author={Liu, Aixin and Feng, Bei and Xue, Bing and Wang, Bingxuan and Wu, Bochao and Lu, Chengda and Zhao, Chenggang and Deng, Chengqi and Zhang, Chenyu and Ruan, Chong and others},
  journal={arXiv preprint arXiv:2412.19437},
  year={2024}
}

@article{liu2024deepseek-V2,
  title={Deepseek-v2: A strong, economical, and efficient mixture-of-experts language model},
  author={Liu, Aixin and Feng, Bei and Wang, Bin and Wang, Bingxuan and Liu, Bo and Zhao, Chenggang and Dengr, Chengqi and Ruan, Chong and Dai, Damai and Guo, Daya and others},
  journal={arXiv preprint arXiv:2405.04434},
  year={2024}
}

@article{mishra2021acceleratingn:m,
  title={Accelerating sparse deep neural networks},
  author={Mishra, Asit and Latorre, Jorge Albericio and Pool, Jeff and Stosic, Darko and Stosic, Dusan and Venkatesh, Ganesh and Yu, Chong and Micikevicius, Paulius},
  journal={arXiv preprint arXiv:2104.08378},
  year={2021}
}

@article{frankle2018lottery,
  title={The lottery ticket hypothesis: Finding sparse, trainable neural networks},
  author={Frankle, Jonathan and Carbin, Michael},
  journal={arXiv preprint arXiv:1803.03635},
  year={2018}
}

@inproceedings{frantar2023sparsegpt,
  title={Sparsegpt: Massive language models can be accurately pruned in one-shot},
  author={Frantar, Elias and Alistarh, Dan},
  booktitle={ICML},
  pages={10323--10337},
  year={2023},
  organization={PMLR}
}

@article{sun2023wanda,
  title={A simple and effective pruning approach for large language models},
  author={Sun, Mingjie and Liu, Zhuang and Bair, Anna and Kolter, J Zico},
  journal={arXiv preprint arXiv:2306.11695},
  year={2023}
}

@article{das2023beyond,
  title={Beyond size: How gradients shape pruning decisions in large language models},
  author={Das, Rocktim Jyoti and Sun, Mingjie and Ma, Liqun and Shen, Zhiqiang},
  journal={arXiv preprint arXiv:2311.04902},
  year={2023}
}

@inproceedings{dong2024pruner,
  title={Pruner-Zero: Evolving Symbolic Pruning Metric From Scratch for Large Language Models},
  author={Dong, Peijie and Li, Lujun and Tang, Zhenheng and Liu, Xiang and Pan, Xinglin and Wang, Qiang and Chu, Xiaowen},
  booktitle={ICML},
  pages={11346--11374},
  year={2024},
  organization={PMLR}
}

@inproceedings{shao2024omniquant,
  title={OmniQuant: Omnidirectionally Calibrated Quantization for Large Language Models},
  author={Shao, Wenqi and Chen, Mengzhao and Zhang, Zhaoyang and Xu, Peng and Zhao, Lirui and Li, Zhiqian and Zhang, Kaipeng and Gao, Peng and Qiao, Yu and Luo, Ping},
  booktitle={ICLR},
  year={2024}
}

@inproceedings{xiao2023smoothquant,
  title={Smoothquant: Accurate and efficient post-training quantization for large language models},
  author={Xiao, Guangxuan and Lin, Ji and Seznec, Mickael and Wu, Hao and Demouth, Julien and Han, Song},
  booktitle={ICML},
  pages={38087--38099},
  year={2023},
  organization={PMLR}
}

@article{zhang2022opt,
  title={Opt: Open pre-trained transformer language models},
  author={Zhang, Susan and Roller, Stephen and Goyal, Naman and Artetxe, Mikel and Chen, Moya and Chen, Shuohui and Dewan, Christopher and Diab, Mona and Li, Xian and Lin, Xi Victoria and others},
  journal={arXiv preprint arXiv:2205.01068},
  year={2022}
}

@article{touvron2023llama,
  title={Llama: Open and efficient foundation language models},
  author={Touvron, Hugo and Lavril, Thibaut and Izacard, Gautier and Martinet, Xavier and Lachaux, Marie-Anne and Lacroix, Timoth{\'e}e and Rozi{\`e}re, Baptiste and Goyal, Naman and Hambro, Eric and Azhar, Faisal and others},
  journal={arXiv preprint arXiv:2302.13971},
  year={2023}
}

@article{touvron2023llama2,
  title={Llama 2: Open foundation and fine-tuned chat models},
  author={Touvron, Hugo and Martin, Louis and Stone, Kevin and Albert, Peter and Almahairi, Amjad and Babaei, Yasmine and Bashlykov, Nikolay and Batra, Soumya and Bhargava, Prajjwal and Bhosale, Shruti and others},
  journal={arXiv preprint arXiv:2307.09288},
  year={2023}
}

@inproceedings{wiki,
  title={Pointer Sentinel Mixture Models},
  author={Merity, Stephen and Xiong, Caiming and Bradbury, James and Socher, Richard},
  booktitle={ICLR},
  year={2017}
}

@article{c4,
  title={Exploring the limits of transfer learning with a unified text-to-text transformer},
  author={Raffel, Colin and Shazeer, Noam and Roberts, Adam and Lee, Katherine and Narang, Sharan and Matena, Michael and Zhou, Yanqi and Li, Wei and Liu, Peter J},
  journal={Journal of machine learning research},
  volume={21},
  number={140},
  pages={1--67},
  year={2020}
}

@inproceedings{clark2019boolq,
  title={BoolQ: Exploring the Surprising Difficulty of Natural Yes/No Questions},
  author={Clark, Christopher and Lee, Kenton and Chang, Ming-Wei and Kwiatkowski, Tom and Collins, Michael and Toutanova, Kristina},
  booktitle={Proceedings of NAACL-HLT},
  pages={2924--2936},
  year={2019}
}

@article{clark2018thinkARC,
  title={Think you have solved question answering? try arc, the ai2 reasoning challenge},
  author={Clark, Peter and Cowhey, Isaac and Etzioni, Oren and Khot, Tushar and Sabharwal, Ashish and Schoenick, Carissa and Tafjord, Oyvind},
  journal={arXiv preprint arXiv:1803.05457},
  year={2018}
}

@inproceedings{zellers2019hellaswag,
  title={HellaSwag: Can a Machine Really Finish Your Sentence?},
  author={Zellers, Rowan and Holtzman, Ari and Bisk, Yonatan and Farhadi, Ali and Choi, Yejin},
  booktitle={ACL},
  pages={4791--4800},
  year={2019}
}

@article{sakaguchi2021winogrande,
  title={Winogrande: An adversarial winograd schema challenge at scale},
  author={Sakaguchi, Keisuke and Bras, Ronan Le and Bhagavatula, Chandra and Choi, Yejin},
  journal={Communications of the ACM},
  volume={64},
  number={9},
  pages={99--106},
  year={2021},
  publisher={ACM New York, NY, USA}
}

@article{mihaylov2018canopenbookQA,
  title={Can a suit of armor conduct electricity? a new dataset for open book question answering},
  author={Mihaylov, Todor and Clark, Peter and Khot, Tushar and Sabharwal, Ashish},
  journal={arXiv preprint arXiv:1809.02789},
  year={2018}
}

@inproceedings{sun2020contrastive,
  title={Contrastive Distillation on Intermediate Representations for Language Model Compression},
  author={Sun, Siqi and Gan, Zhe and Fang, Yuwei and Cheng, Yu and Wang, Shuohang and Liu, Jingjing},
  booktitle={EMNLP},
  pages={498--508},
  year={2020}
}

@inproceedings{sun2019patient,
  title={Patient Knowledge Distillation for BERT Model Compression},
  author={Sun, Siqi and Cheng, Yu and Gan, Zhe and Liu, Jingjing},
  booktitle={EMNLP},
  pages={4323--4332},
  year={2019}
}

@article{sajjad2020poor,
  title={Poor man’s bert: Smaller and faster transformer models},
  author={Sajjad, Hassan and Dalvi, Fahim and Durrani, Nadir and Nakov, Preslav},
  journal={arXiv preprint arXiv:2004.03844},
  volume={2},
  number={2},
  year={2020}
}

@inproceedings{liu2023deja,
  title={Deja vu: Contextual sparsity for efficient llms at inference time},
  author={Liu, Zichang and Wang, Jue and Dao, Tri and Zhou, Tianyi and Yuan, Binhang and Song, Zhao and Shrivastava, Anshumali and Zhang, Ce and Tian, Yuandong and Re, Christopher and others},
  booktitle={ICML},
  pages={22137--22176},
  year={2023},
  organization={PMLR}
}

@article{li2024repquant,
  title={Repquant: Towards accurate post-training quantization of large transformer models via scale reparameterization},
  author={Li, Zhikai and Liu, Xuewen and Zhang, Jing and Gu, Qingyi},
  journal={arXiv preprint arXiv:2402.05628},
  year={2024}
}

@article{ma2023llm,
  title={Llm-pruner: On the structural pruning of large language models},
  author={Ma, Xinyin and Fang, Gongfan and Wang, Xinchao},
  journal={Advances in neural information processing systems},
  volume={36},
  pages={21702--21720},
  year={2023}
}

\end{document}